\pdfoutput=1

\documentclass[11pt]{article}
\usepackage{amsmath}  
\usepackage{amssymb}  

\usepackage[final]{acl}

\usepackage{times}
\usepackage{latexsym}

\usepackage[T1]{fontenc}

\usepackage[utf8]{inputenc}
\usepackage{mathtools}
\usepackage{microtype}

\usepackage{inconsolata}

\usepackage{graphicx}

\usepackage{diagbox}
\usepackage{url}
\usepackage{subcaption}
\usepackage{float}
\usepackage{multirow}
\usepackage{tcolorbox}
\usepackage{booktabs}
\usepackage{adjustbox}
\usepackage{enumitem}
\usepackage{caption}
\usepackage{shadow}
\usepackage{ulem}
\usepackage{longtable}
\usepackage{array}
\usepackage{algpseudocode}

\usepackage{booktabs}
\usepackage{colortbl}
\usepackage{arydshln}
\usepackage{algorithm}
\usepackage{listings}
\title{From ``Aha Moments'' to Controllable Thinking: Toward Meta-Cognitive Reasoning in LRMs via Decoupled Reasoning and Control}

\author{
    \textbf{Rui Ha\textsuperscript{1}},
    \textbf{Rui Pu\textsuperscript{1}},
    \textbf{Chaozhuo Li\textsuperscript{1}},
    \textbf{Li Sun\textsuperscript{1}},
    \textbf{Sen Su\textsuperscript{1,2,\dag}}
    \\
    \\
    \textsuperscript{1}Beijing University of Posts and Telecommunications, China \\
    \textsuperscript{2}Chongqing University of Posts and Telecommunications, China \\
    \{harry, puruirui, lichaozhuo, lsun, susen\}@bupt.edu.cn
    \\
}

\usepackage{amsmath}
\usepackage{caption}
\usepackage{enumitem}
\usepackage{enumerate}
\usepackage{algorithm}
\usepackage{algorithmicx}
\usepackage{algpseudocode}
\usepackage{graphicx}
\usepackage{multirow} 
\usepackage{booktabs} 
\usepackage{mdframed}
\usepackage{supertabular}
\usepackage{tikz}
\usepackage{hyperref}
\usepackage{pdfcomment}
\usepackage{xcolor}

\begin{document}
\maketitle
\renewcommand{\thefootnote}{\fnsymbol{footnote}} 
\footnotetext[2]{The corresponding author.}  
\renewcommand{\thefootnote}{\arabic{footnote}}
\begin{abstract} 
Large Reasoning Models (LRMs) can exhibit step-by-step reasoning, reflection, and backtracking, but these behaviors are often unregulated, leading to overthinking. As a result, LRMs continue generating redundant reasoning even after reaching high-confidence conclusions. This increases inference cost and latency, limiting practical deployment. The root cause is the absence of an intrinsic mechanism to monitor the reasoning state and decide when to continue, backtrack, or stop. We propose MERA, a meta-cognitive reasoning framework that decouples reasoning from control to enable independent optimization of control strategies. MERA constructs high-quality reasoning–control supervision data via a takeover-based pipeline, and transforms long-horizon traces into structured reasoning–control alternating sequences for training. The model is trained with supervised fine-tuning to internalize the structured separation, and further optimized with Control-Segment Policy Optimization (CSPO), which combines segment-wise GRPO with control masking to focus learning on control segments. Experiments across reasoning benchmarks show that MERA improves both efficiency and accuracy.

\end{abstract}

\section{Introduction}   

\begin{figure}[ht!]
\centering
\includegraphics[width=\linewidth]{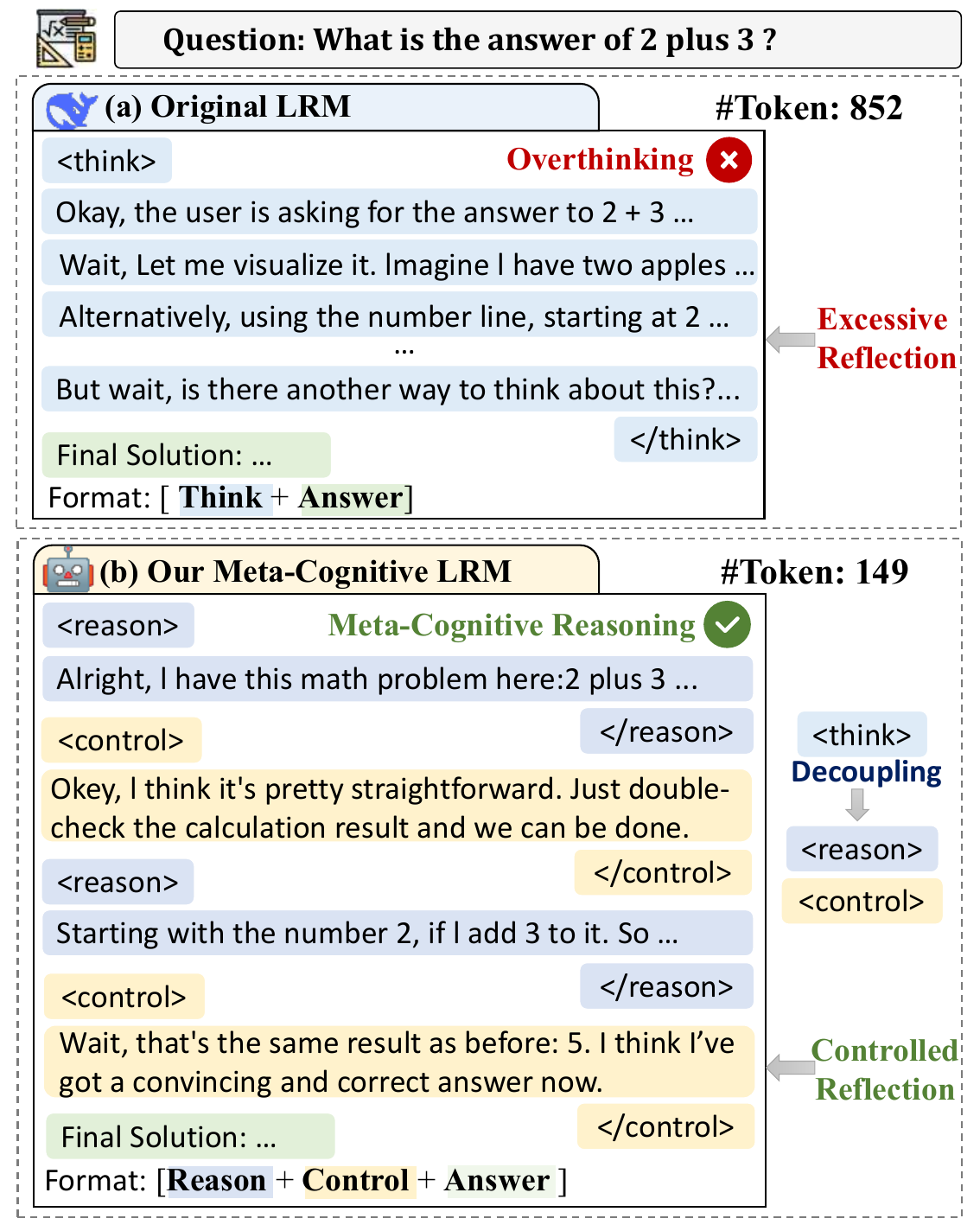}
\captionsetup{justification=justified,singlelinecheck=false}
\caption{The comparison of reasoning answers between Original LRM and our proposed  Meta-Cognitive LRM.}
\label{fig:intro}
\end{figure}

Large Reasoning Models (LRMs) have achieved significant advancements in complex tasks such as mathematical problem-solving and symbolic reasoning by integrating cognitive operations including step-by-step reasoning, reflection, and backtracking~\cite{LRM1,LRM2,wang2026devil}. These emergent capabilities, often described as “Aha Moments,” demonstrate the model's latent ability to engage in complex test-time reasoning~\cite{guo2025deepseek,li2025loki}. However, such cognitive behaviors remain unregulated and uncontrolled, frequently resulting in overthinking, where the model continues generating redundant reasoning content even after reaching high-confidence conclusions~\cite{overthinking,wu2025lapo,zhao2022learning}. This leads to substantial computational overhead and increased response latency, thereby limiting the practical deployment of LRMs~\cite{arora2025training,mrt,liu2026clawkeeper}.

The current approaches to addressing this challenge can be classified into three main categories. Methods that directly shorten the reasoning length often lack adaptability to problems of varying difficulty, which may result in excessive shortening and a subsequent decline in performance~\cite{o1-pruner, nowait}. Some methods introduce preset budget mechanisms, which either rely on coarse-grained control between fast and slow thinking modes~\cite{zhang2025adaptthink, lou2025adacot}, or impose length constraints prior to answering~\cite{dast, aggarwal2025l1}, both lacking the flexibility to adjust the model's state during the reasoning process. Dynamic early-stopping methods, while capable of determining the optimal termination point of reasoning, still fundamentally rely on external evaluation metrics such as confidence thresholds to make decisions~\cite{yang2025dynamic, qiao2025concise}.

The limitation of these approaches lies in their treatment of length control as an external intervention. 
Specifically, they rely on predefined rules or external metrics, rather than enabling the model to decide when to stop based on its own reasoning state. 
In contrast, humans can flexibly allocate cognitive resources in real time during problem solving~\cite{mathcog}. 
This leads to a critical question: \textit{Are LRMs capable of effectively regulating their own behavior during reasoning?}

As shown in Figure~\ref{fig:intro}(a), even when reliable intermediate conclusions have already been reached, LRMs tend to repeatedly perform cognitive operations such as reflection and backtracking, resulting in substantial redundant generation. 
This indicates that LRMs lack an internal self-regulation mechanism during reasoning: they often struggle to decide when to continue, whether reflection or backtracking is warranted, or when to terminate reasoning.
Inspired by the concept of meta-cognition in cognitive psychology~\cite{cog1,liu2025scales}, which refers to the awareness and regulation of one's own cognitive processes, we characterize this issue as a deficiency in the model’s meta-cognitive capacity. 

In current frameworks, control behaviors are often entangled with the reasoning process and jointly optimized under the same objective function. Control behaviors often degrade into static strategies, and models trained under this framework prioritize generating correct answers rather than optimizing the reasoning path, further exacerbating redundant generation in the reasoning process.

To address the above issue, we propose the Meta-cognitive Reasoning Framework (MERA). Unlike conventional methods that treat the thinking process as a unified whole, MERA’s core innovation lies in decoupling the thinking process within LRMs into two distinct components: reasoning and control. By employing structurally separated training and optimization, MERA facilitates authentic meta-cognitive regulation. As illustrated in Figure~\ref{fig:intro}(b), unlike prior models that blindly repeat cognitive operations without self-awareness, the models trained with MERA first assesse the current reasoning state. Upon recognizing that the reasoning output is accurate and reliable, they promptly issue a definitive control signal to terminate the reasoning process.

However, introducing an independent control component brings three new challenges.
First, there is a severe scarcity of reasoning-control data, as existing training datasets generally lack high-quality meta-cognitive control annotations, and manually labeling such fine-grained data is extremely costly.
Second, distinguishing control behaviors from reasoning content presents a notable challenge, as the reasoning traces in LRMs are often lengthy and lack explicit structural boundaries, making it difficult to clearly separate the two.
Third, control instructions are often dispersed across multiple segments of reasoning, which causes policy optimization signals to be diluted by non-critical content during training.

To address these challenges, our framework introduces three key mechanisms.
First, to mitigate the scarcity of high-quality reasoning-control data, a control-takeover mechanism is designed, which identifies critical moments during reasoning where meta-cognitive control is required, and delegates control generation at these points to auxiliary LLMs.
Second, a structured decoupling mechanism is implemented via supervised fine-tuning, enabling the model to generate explicit reasoning and control tags, thereby achieving clear separation between reasoning and control processes.
Third, a Control-Segment Policy Optimization (CSPO) method is proposed, which combines segment-wise GRPO and control masking to enable targeted optimization of control behavior with minimal interference from irrelevant content.

This paper makes the following contributions:
\begin{itemize}
  \item We frame overthinking as a deficiency in fine-grained internal control, and we propose a reasoning-control decoupling strategy to endow LRMs with meta-cognitive regulatory capabilities.
  \item We propose MERA, a framework that integrates control-takeover, structural separation, and CSPO to overcome key challenges in independent control optimization.
  \item Comprehensive experiments show that MERA significantly improves both accuracy and efficiency on various reasoning benchmarks.
\end{itemize}

\begin{figure*}[ht!]
\centering
\includegraphics[width=\linewidth]{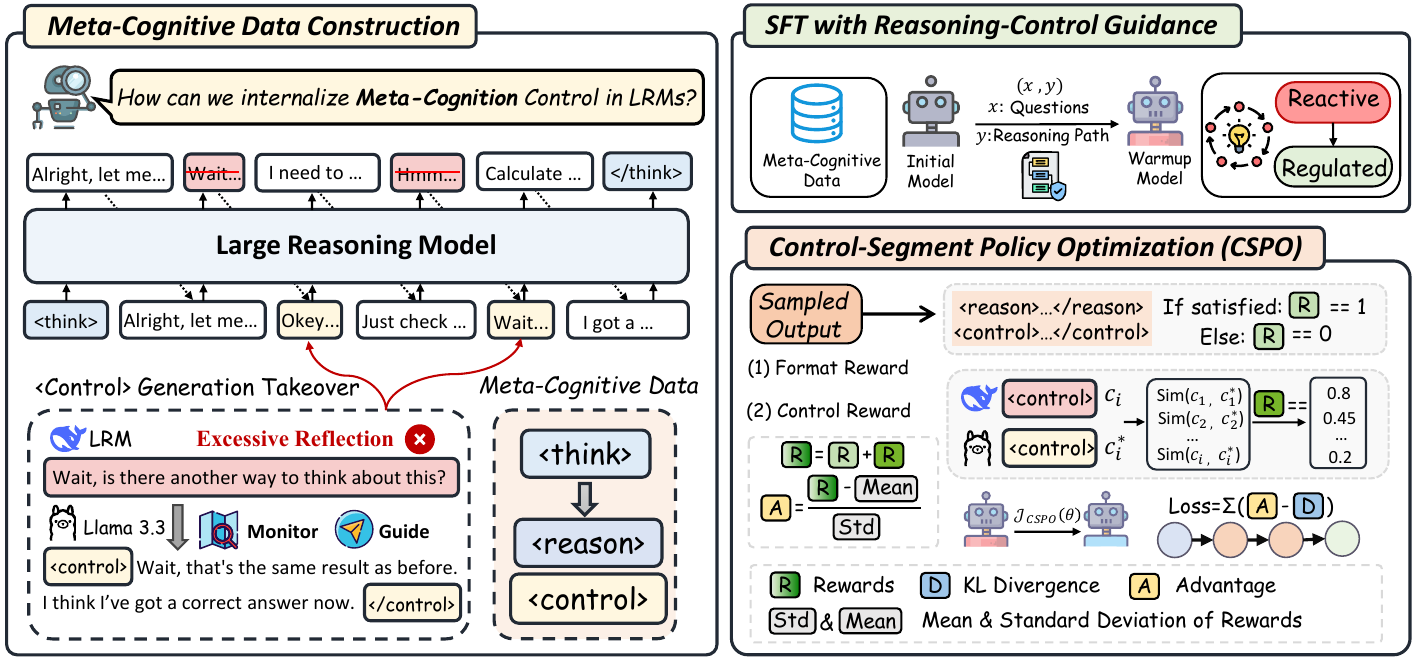}
\captionsetup{justification=justified,singlelinecheck=false}
\caption{Overview of the proposed Meta-cognitive Reasoning Framework (MERA). The framework consists of three key components: Meta-cognitive Data Construction, SFT with Reasoning-Control Guidance, and Control-Segment Policy Optimization. }
\label{fig:framework}
\end{figure*}
\section{Methodology}
To address the issue of overthinking exhibited by LRMs in complex tasks, we propose the \textbf{Meta-cognitive Reasoning Framework (MERA)}.
This architecture injects structured, self-regulatory meta-cognitive capabilities into LRMs.
As illustrated in Figure~\ref{fig:framework}, MERA comprises three interrelated components: a decoupled modeling mechanism that separates reasoning and control, a control-driven data construction pipeline, and a training paradigm that integrates supervised fine-tuning with CSPO. 
Together, these modules enable the model to explicitly monitor, evaluate, and regulate its internal reasoning processes.

\subsection{Decoupled Modeling of Thought Processes}
\subsubsection{Decoupling Definition}
Traditional LRMs typically treat the entire reasoning process as a monolithic text generation stream, lacking built-in mechanisms for introspection and self-regulation. To enable finer-grained control over the internal cognitive process, we propose a structural decoupling of \textbf{reasoning} and \textbf{control} during generation. Specifically, we decompose the overall cognitive process into the following two functional modules: 
\begin{itemize}
    \item $r_k \in \mathcal{R}$: a \textbf{reasoning statement}, which represents a logical expression or step used to solve the task.
    \item $c_k \in \mathcal{C}$: a \textbf{control statement}, responsible for assessing and regulating the reasoning process.
\end{itemize}
The model output is composed of an alternating sequence of reasoning and control segments, formalized as:
\begin{equation}
\tau = \left\{(r_1, c_1),\ (r_2, c_2),\ \dots,\ (r_K, c_K)\right\}.
\end{equation}

Here, each pair $(r_k, c_k)$ represents the $k$-th round of reasoning–control interaction.

Reasoning segments are denoted using the \texttt{\textless reason\textgreater} tag, while control segments are marked with the \texttt{\textless control\textgreater} tag. In line with prior work~\cite{mathod1}, transitions between reasoning and control commonly occur at \textit{cognitive turning points}, such as ``wait'', ``hmm'', or ``alternative'', which act as natural linguistic cues that signal the need for meta-cognitive intervention.

\subsubsection{Problem Formalization}
Under this structural framework, we formalize the meta-cognitive reasoning task as a conditional generation problem. Given an input query $x$, the model is required to generate an alternating reasoning–control sequence $\tau$, followed by the final answer $y$. The overall generation probability can be factorized as:
\begin{equation}
\pi_\theta(\tau, y \mid x) = \pi_\theta(y \mid \tau, x) \cdot \pi_\theta(\tau \mid x),
\end{equation}
where $\pi_\theta(\tau \mid x)$ denotes the probability of generating the structured reasoning–control sequence conditioned on the input, and $\pi_\theta(y \mid \tau, x)$ represents the probability of producing the final answer based on both the input and the generated reasoning trajectory.

\subsection{Meta-cognitive Data Construction}

To develop reasoning models with autonomous regulation capabilities, we construct a dataset enriched with explicit meta-cognitive signals, enabling the model not only to generate reasoning traces but also to dynamically monitor and control the reasoning process. The data construction process is divided into three structured stages: identification of control takeover points, generation of control signals, and construction of alternating sequences.

\subsubsection{Identification of Control Takeover}
We design a control takeover mechanism to automatically identify key takeover points in LRM reasoning. Long-horizon traces often exhibit step-by-step reasoning, reflection, and backtracking, which naturally form discrete segments. Segment transitions are typically marked by turning expressions such as ``wait'' and``alternatively'', indicating hesitation, self-reflection, or shifts in reasoning. We treat these markers as control takeover signals, grounded in prior findings that such turning expressions consistently align with reasoning-state transitions and can serve as reliable anchors for trace segmentation~\cite{yang2025dynamic,overthinking,NIPS1}. Using these anchors, we insert explicit control instructions to enable a structured separation of reasoning and control.

\subsubsection{Generation of Control Signals}
After locating the takeover points, we use the Llama-3.3-70B-Instruct model~\cite{llama3} to generate key control statements. Specifically, we design structured prompt templates that simulate a ``meta-cognitive monitor'' observing the model's reasoning process and request the following two tasks: evaluating the current reasoning and providing control suggestions. Based on the assessment of the current reasoning state, the model generates control statements and inserts them after a reasoning segment.

\subsubsection{Construction of Alternating Sequences}
After generating the control statements, we return the generation phase to the original LRMs to continue producing subsequent content. This process ensures that the reasoning chain alternates naturally between ``control intervention–continue thinking'' in a structured manner. Finally, we integrate the complete reasoning trajectory with the generated control statements, constructing the \textbf{reasoning–control alternating sequence} for model training. Each training sample also includes the final answer, allowing the model to simultaneously learn process regulation and task completion. The entire process automatically transforms the original reasoning data into structured samples in the form of triples $(x_i, \tau_i, y_i)$, where $x_i$ is the input query, $\tau_i$ is the alternating reasoning–control sequence, and $y_i$ is the final answer.

\subsection{SFT with Reasoning-Control Guidance}
To effectively guide the model in mastering the structured reasoning–control pattern, we design a unified supervised fine-tuning (SFT) mechanism based on the constructed dataset, allowing the model to simultaneously learn incremental reasoning and meta-cognitive regulation abilities during the generation process.

\subsubsection{Joint Generation Modeling}
During the training phase, each sample is modeled as a triplet $(x_i, \tau_i, y_i)$, where $x_i$ represents the input query, $\tau_i$ is the structured intermediate process formed by alternating reasoning and control segments, and $y_i$ is the final answer. We adopt the standard conditional language modeling objective to model the answer path:
\begin{equation}
\mathcal{L}_{\text{SFT}}(\theta) = - \sum_{i=1}^{N} \log \pi_\theta\left(\tau_i, y_i \mid x_i\right).
\end{equation}

After the initial SFT, the model acquires the ability to decouple the original reasoning process by appropriately using the \texttt{<reason>} and \texttt{<control>} tags. It also learns to generate effective control content that includes both evaluative feedback and directive signals to guide the reasoning.

\subsection{Control-Segment Policy Optimization}
To further enhance the self-regulatory capabilities of LRMs during reasoning, we propose \textbf{Control Segment Policy Optimization (CSPO)}, a training framework designed to address two major challenges in reinforcement learning. First, control directives are often distributed across multiple reasoning segments, making it difficult for standard GRPO to attribute rewards with fine granularity; Second, control content is highly sparse within sequences, causing policy updates to be easily diluted by non-critical positions. CSPO addresses these issues through the following three mechanisms:
\textbf{(1) Segmented GRPO Modeling}, which provides independent reward feedback for each reasoning–control unit; 
\textbf{(2) Control Reward Modeling}, combining semantic and structural signals to guide learning; 
\textbf{(3) Control Masking}, which restricts optimization to control-relevant tokens, thus improving learning efficiency and stability.

\subsubsection{Segmented GRPO Modeling}
Inspired by prior work~\cite{mathod4}, we partition the generated sequence into multiple segments ($\tau_i$) to capture fine-grained control features within each segment. For each segment, we sample an output $o_i$ from the policy and independently evaluate its corresponding reward. We adopt the GRPO policy optimization method~\cite{mathod5}, wherein $G$ complete outputs are sampled from the previous policy $\pi_{\theta_{\text{old}}}$, followed by reward normalization and computation of the advantage function for each segment:

\begin{equation}
\hat{A}_{i,t} = \frac{r(o_i) - \text{mean}(r(o_1), \dots, r(o_G)\})}{\text{std}(\{r(o_1), \dots, r(o_G)\})}.
\end{equation}

Here, $r(o_i)$ is the segment-level reward function defined below, computed from independent feedback per segment.

\subsubsection{Control Reward Modeling}
We design a control reward function $r(o_i)$ composed of two complementary components that evaluate both semantic correctness and structural conformity of the generated control content:

\textbf{Control Reward ($R_{\text{ctrl}}$):} This term measures whether the generated control segment ${c}_k$ semantically aligns with the reference control target $c_k^*$. We compute semantic similarity using the GPT-4o model~\cite{gpt4}:
\begin{equation}
R_{\text{ctrl}}(c_k) = \text{Score}_{\text{Sim}}(c_k, c_k^*).
\end{equation}

\textbf{Format Reward ($R_{\text{format}}$):} This term encourages the model to follow a standardized structure, particularly the \texttt{<reason>} and \texttt{<control>} format, which improves structural consistency and interpretability. The overall reward for each segment is computed as:
\begin{equation}
r(o_i) = R_{\text{ctrl}} + R_{\text{format}}.
\end{equation}

This mask ensures that only tokens within control spans receive gradient updates, which significantly improves the precision and efficiency of policy optimization in control-sensitive tasks. The final objective function for CSPO is:
\begin{equation}\small
\begin{split}
& \mathcal{J}_{\text{CSPO}}(\theta) = \mathbb{E}_{x, o} 
\frac{1}{Z} \sum_{k=1}^{K}  \frac{1}{G} \sum_{i=1}^{G}\frac{1}{|o_i|} \sum_{t=1}^{|o_i|}
\Bigg[ M_k \min  \Big( r_t(\theta)  \\
&\hat{A}_{i,t} , 
\mathrm{clip}(r_t(\theta), 1{-}\epsilon, 1{+}\epsilon) \hat{A}_{i,t} \Big) 
- \beta D_{\text{KL}}(\pi_\theta \| \pi_{\text{ref}}) 
\Bigg]
\end{split},
\end{equation}
where $r_t(\theta)$ denotes the policy ratio at time step $t$, $D_{\text{KL}}$ penalizes the divergence from the reference policy $\pi_{\text{ref}}$ and $Z = \sum_{k=1}^{K} M_k$ is the normalization factor. This objective ensures that optimization is exclusively directed at control tokens, with high-quality structural and semantic signals driving stable and efficient policy learning.

\begin{table*}[ht]
\setlength{\tabcolsep}{1pt} 
\resizebox{\textwidth}{!}{
\belowrulesep=0pt
\aboverulesep=0pt
\footnotesize
\renewcommand{\arraystretch}{1}
\begin{tabular}{@{}lccccccccccll@{}} 
\toprule
 \multirow{2}{*}{\textbf{Method}} 
 & \multicolumn{2}{c}{\textbf{GSM8K}} & \multicolumn{2}{c}{\textbf{MATH-500}} & \multicolumn{2}{c}{\textbf{AMC 2023}} & \multicolumn{2}{c}{\textbf{AIME 2024}} & \multicolumn{2}{c}{\textbf{AIME 2025}}    & \multicolumn{2}{c}{\textbf{Overall}} \\ 
    & {Acc$\uparrow$} & {Tokens$\downarrow$} & {Acc$\uparrow$} & {Tokens$\downarrow$} & {Acc$\uparrow$}  & {Tokens$\downarrow$} & {Acc$\uparrow$} & {Tokens$\downarrow$} & {Acc$\uparrow$} & {Tokens$\downarrow$}  & {Acc$\uparrow$} & {Tokens$\downarrow$} \\ 
\midrule
\multicolumn{13}{c}{\textit{DeepSeek-R1-Distill-Qwen-1.5B}} \\ \cmidrule{1-13}
\textit{Original} & 86.1 & 2,245 & 83.0 & 3,978 & 67.7 & 7,160 & 29.3 & 13,832 & 26.9 & 14,680 & 58.6 & 8,379 \\ 
\hdashline[1pt/2pt]
\addlinespace[2pt]
\textit{O1-Pruner} & 85.1 & 1,535 & 82.3 & 2,446 & 69.5 & 5,622 & 27.5 & 12,155 & 24.1 & 12,701 & 57.7 & 6,892 \\ 

\textit{No Wait} & 84.9 & 1,955 & 81.6 & 2,894 & 68.5 & 6,422 & 26.1 & 9,167 & 20.3 & 11,601 & 56.3 & 6,408 \\ 

\textit{DAST} & 85.6 & 1,783 & 83.4 & 3,155 & 70.2 & 5,583 & 29.5 & 10,042 & 20.6 & 10,647 & 57.9 & 6,242 \\ 

\textit{FCS+Ref.} & 87.7 & 1,397 & 82.9 & 2,883 & 72.1 & 4,449 & 29.7 & 9,898 & 28.6 & 11,624 & 60.2 & 6,050 \\ 

\textit{LCPO} & 83.5 & 1,890 & 80.5 & 2,684 & 63.9 & 6,694 & 23.5 & 9,692 & 24.7 & 10,075 & 55.2 & 6,207 \\ 

\textit{DEER} & 86.2 & 1,205 & 84.1 & 2,398 & 70.2 & 4,179 & 26.6 & 9,732 & 20.2 & 9,890 & 57.5 & 5,481 \\ 

\midrule
\rowcolor[cmyk]{0.017,0.017,0.017,0.017}\textbf{MERA(Ours)} & \textbf{89.3} & \textbf{1,108} & \textbf{86.0} & \textbf{2,239} & \textbf{74.5} & \textbf{3,180} & \textbf{31.2} & \textbf{8,425} & \textbf{30.9} & \textbf{9,357} & \textbf{62.4} & \textbf{4,862} \\ 
\midrule

\multicolumn{13}{c}{\textit{DeepSeek-R1-Distill-Qwen-7B}} \\ \cmidrule{1-13}
\textit{Original} & 90.2 & 1,819 & 86.9 & 3,422 & 77.4 & 6,738 & 53.1 & 12,185 & 48.2 & 13,276 & 71.2 & 7,488 \\ 
\hdashline[1pt/2pt]
\addlinespace[2pt]

\textit{O1-Pruner} & 92.5 & 1,052 & 89.0 & 2,678 & 82.8 & 7,501 & 52.2 & 9,412 & 49.4 & 10,973 & 73.2 & 6,323 \\ 

\textit{No Wait} & 89.8 & 1,733 & 87.2 & 2,579 & 75.7 & 5,478 & 42.5 & 10,048 & 35.2 & 11,827 & 66.1 & 6,333 \\

\textit{DAST} & 92.7 & 1,558 & 88.6 & 2,876 & 80.3 & 4,601 & 52.6 & 10,240 & 49.5 & 9,721 & 72.7 & 5,799 \\

\textit{FCS+Ref.} & 92.9 & 1,218 & 87.8 & 2,909 & 81.5 & 5,143 & 55.1 & 9,212 & 49.8 & 9,844 & 73.4 & 5,665 \\

\textit{LCPO} & 88.0 & 1,488 & 84.7 & 2,539 & 73.6 & 4,821 & 45.3 & 9,408 & 40.6 & 9,904 & 66.4 & 5,632 \\ 

\textit{DEER} & 89.0 & 1,082 & 88.9 & 1,908 & 82.2 & 5,194 & 46.4 & 9,932 & 39.6 & 9,302 & 69.2 & 5,484 \\ 
\midrule
\rowcolor[cmyk]{0.017,0.017,0.017,0.017}\textbf{MERA(Ours)} & \textbf{93.7} & \textbf{822} & \textbf{91.0} & \textbf{1,739} & \textbf{85.7} & \textbf{3,711} & \textbf{56.1} & \textbf{8,398} & \textbf{53.6} & \textbf{8,732} & \textbf{76.0} & \textbf{4,680} \\ 
\midrule

\multicolumn{13}{c}{\textit{DeepSeek-R1-Distill-Qwen-14B}} \\ \cmidrule{1-13}
\textit{Original} & 92.3 & 1,917 & 87.4 & 2,832 & 83.4 & 7,925 & 60.6 & 11,155 & 56.4 & 12,752 & 76.0 & 7,316 \\ 
\hdashline[1pt/2pt]
\addlinespace[2pt]

\textit{O1-Pruner} & 91.9 & 1,263 & 90.2 & 2,015 & 84.2 & 7,381 & 55.2 & 9,003 & 53.1 & 9,356 & 74.9 & 5,804 \\ 

\textit{No Wait} & 92.6& 1,152 & 88.7 & 1,931 & 79.6 & 5,709 & 53.8 & 9,692 & 45.3 & 8,280 & 72.0 & 5,353 \\

\textit{DAST} & 94.8 & 1,778 & 88.9 & 1,943 & 84.9 & 4,338 & 55.6 & 7,768 & 56.2 & 10,298 & 76.1 & 5,225 \\

\textit{FCS+Ref.} & 94.7 & 1,141 & 89.3 & 1,937 & 83.0 & 5,281 & 56.4 & 7,749 & 53.7 & 8,309 & 75.4 & 4,883 \\

\textit{LCPO} & 91.2 & 1,645 & 86.8 & 2,426 & 78.1 & 5,533 & 56.1 & 9,460 & 50.2 & 10,405 & 72.5 & 5,894 \\ 

\textit{DEER} & 93.5 & 975 & 87.7 & 1,729 & 77.2 & 3,291 & 60.7 & 8,388 & 47.3 & 9,723 & 73.3 & 4,821 \\ 
\midrule
\rowcolor[cmyk]{0.017,0.017,0.017,0.017}\textbf{MERA(Ours)} & \textbf{96.1} & \textbf{835} & \textbf{92.5} & \textbf{1,359} & \textbf{88.0} & \textbf{3,109} & \textbf{63.3} & \textbf{6,695} & \textbf{59.2} & \textbf{7,320} & \textbf{79.8} & \textbf{3,864} \\ 
\bottomrule
\end{tabular}}
\caption{Performance comparison of MERA and baseline methods on five mathematical reasoning benchmarks.} 
\label{tab1}
\end{table*}

\section{Experiment}
\newcommand{\annotate}[3]{%
    #1\raisebox{-0.5ex}{\scriptsize\textcolor{#2}{#3}}%
}
\subsection{Experimental Setup}
\noindent \textbf{Training datasets.} 
We construct our training set using approximately 5,000 question–answer pairs selected from the DeepScaleR Preview-Dataset~\cite{luo2025deepscaler}. This dataset is a challenging collection of mathematics problems, covering a wide range of difficulty levels and drawing from sources such as AIME (1984–2023), AMC (prior to 2023), and the MATH training set. Consequently, the resulting training set has no overlap with our benchmark dataset.

\noindent \textbf{Evaluation Datasets.}
We conduct comprehensive evaluations of our method on several widely recognized benchmarks for mathematical reasoning. Specifically, we assess performance on the test sets of GSM8K~\cite{data1}, MATH-500~\cite{data3}, AMC2023, as well as AIME2024 and AIME2025~\cite{data8}. These benchmarks cover a diverse range of problem types and difficulty levels, providing a rigorous testbed for evaluating mathematical reasoning capabilities. Additionally, we evaluate our models on MMLU-Pro~\cite{data9} to assess their generalization beyond the mathematical domain, selecting 100 questions from each of three different fields.

\noindent \textbf{Base Models.} 
We adopt the open-source DeepSeek-R1-Distill model series~\cite{guo2025deepseek}, including DeepSeek-R1-Distill-Qwen-1.5B, DeepSeek-R1-Distill-Qwen-7B, and DeepSeek-R1-Distill-Qwen-14B, as our base models. These models are obtained through supervised fine-tuning on reasoning data generated by the DeepSeek-R1 model.

\noindent \textbf{Baselines.}
The baselines include in our comparison fall into three distinct categories:(1) Methods that directly reduce reasoning length, including O1-Pruner~\cite{o1-pruner} and No Wait~\cite{nowait} ; (2) Methods that rely on preset computational budgets before inference, including DAST~\cite{dast}, FCS+Ref.~\cite{overthinking} and LCPO~\cite{aggarwal2025l1}; (3) Methods that dynamically determine termination, including DEER~\cite{yang2025dynamic}.

\noindent \textbf{Evaluation Metrics.} 
We evaluate the proposed method using two key metrics: Accuracy(ACC) and Generation Length(Tokens). ACC is calculated as \(\text{ACC} = \frac{1}{N} \sum_{i=1}^N \mathbb{I}\{\mathcal{M}(\mathcal{LLM}(x_i)) = y_i\}\), where \(x_i\) is the input question, \(y_i\) is the ground-truth answer, \(\mathcal{LLM}(\cdot)\) denotes the model’s output, \(\mathcal{M}(\cdot)\) extracts the predicted answer according to a predefined format (e.g., starting with “The answer is…”). Tokens measures the average generated tokens, computed as \(\text{Tokens} = \frac{1}{N} \sum_{i=1}^N |\mathcal{LLM}(x_i)|\), where \(|\cdot|\) counts the tokens of generated words.

\noindent \textbf{Implementation Details.}
All experiments are conducted using the \texttt{vLLM} framework under a zero-shot chain-of-thought (CoT) setting. The prompt used is: \textit{``Please reason step by step and put the final answer in \textbackslash boxed\{\}."} 
To ensure reliability and statistical validity, each model and configuration is evaluated across five sampling runs.
To ensure complete reasoning traces are captured, the maximum number of new tokens is set to 16384 for AIME and 8192 for all other datasets. 
\begin{figure}
	\centering
	\includegraphics[width=0.95\linewidth]{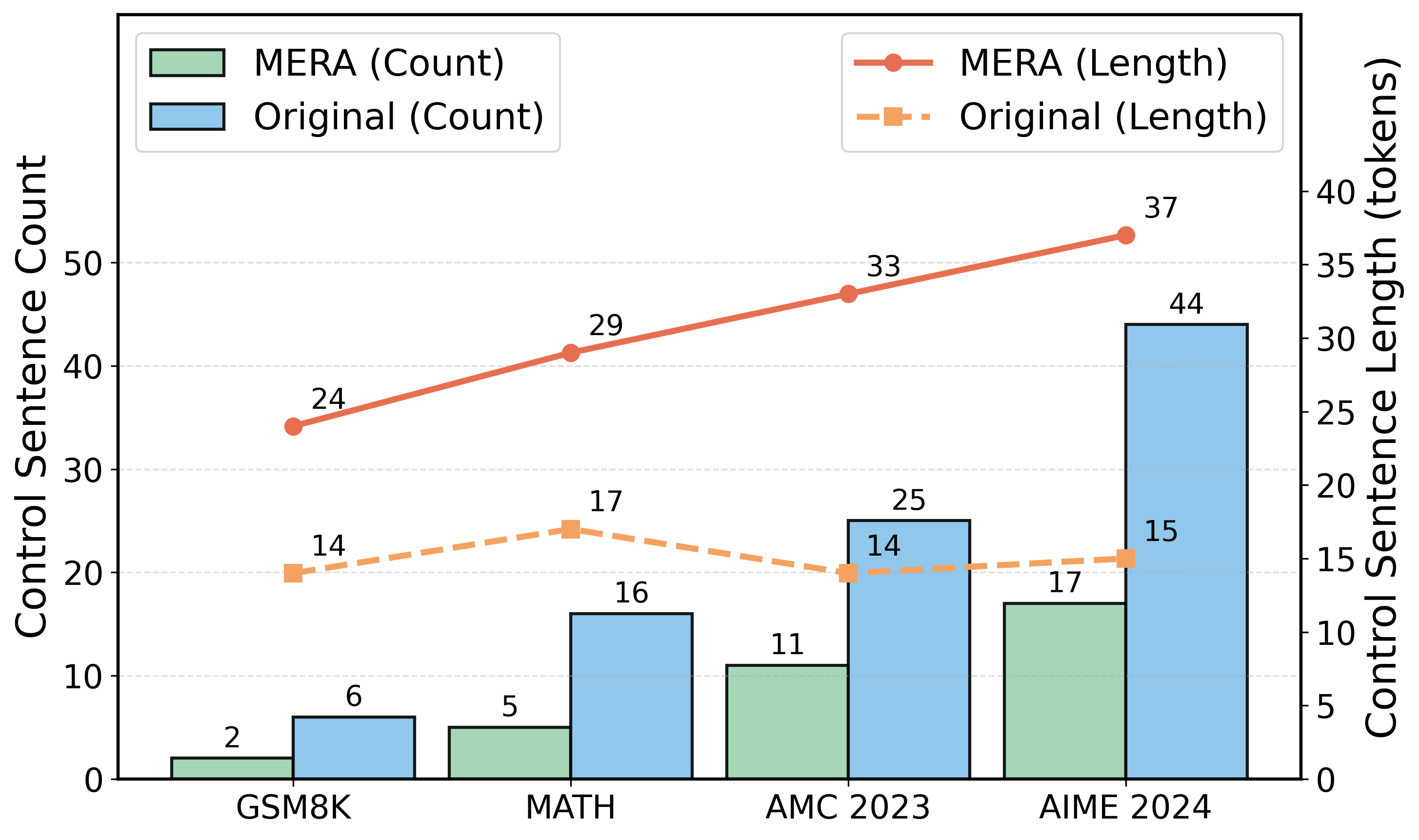}
        \caption{Analysis of control sentence before and after the application of MERA on DeepSeek-R1-Distill-Qwen-7B.}

	\label{fig:control}
\end{figure}

\begin{table}[t!]
\centering
\setlength{\tabcolsep}{0.5pt}  
\begin{tabular}{l|cc|cc|cc}
\toprule
& \multicolumn{2}{c|}{\textbf{Law}} & \multicolumn{2}{c|}{\textbf{Engineering}} & \multicolumn{2}{c}{\textbf{Physics}} \\
\textbf{Method} & \textbf{ACC} & \textbf{LEN} & \textbf{ACC} & \textbf{LEN} & \textbf{ACC} & \textbf{LEN} \\
\midrule
Original & 23.5 & 1528 & 26.4 & 5412 & 34.1 & 4978 \\
O1-Pruner & 22.8 & 1054 & 25.6 & 4981 & 35.6 & 4587 \\
DAST  & 21.4 & 1249 & 27.1 & 5821 & 33.5 & 5243 \\
FCS+Ref.  & 24.1 & 973 & 27.9 & 5518 & 34.9 & 5098 \\
\midrule
MERA (Ours) & \textbf{25.7} & \textbf{814} & \textbf{28.5} & \textbf{4092} & \textbf{36.9} & \textbf{3854} \\
\bottomrule
\end{tabular}
\caption{Evaluation on MMLU-Pro to validate generalization using DeepSeek-R1-Distill-Qwen-7B.}
\label{tbl:diff_domains}
\end{table}

\subsection{Experimental Results}

\subsubsection{Main Results.}
Table~\ref{tab1} presents the performance of MERA across five reasoning benchmarks, evaluated by answer accuracy (Acc) and generation length (Tokens). Compared to existing methods, MERA achieves higher accuracy while significantly reducing reasoning length across all datasets. For instance, on the DeepSeek-R1-Distill-Qwen-1.5B model, MERA reduces the average length from 8,379 tokens to 4,862, while improving accuracy from 58.6\% to 62.4\%. In contrast, MERA not only reduces reasoning length but also improves accuracy, achieving the best efficiency overall. These improvements are attributed to MERA’s explicit enablement of self-regulation during reasoning, which allows the model to dynamically perceive its current cognitive state and perform fine-grained, stage-wise control over the reasoning trajectory. These results demonstrate that structured meta-cognition effectively mitigates overthinking and enhances the model's efficiency in handling problems of varying complexity.

\begin{figure}[tb]
	\centering
	\includegraphics[width=\linewidth]{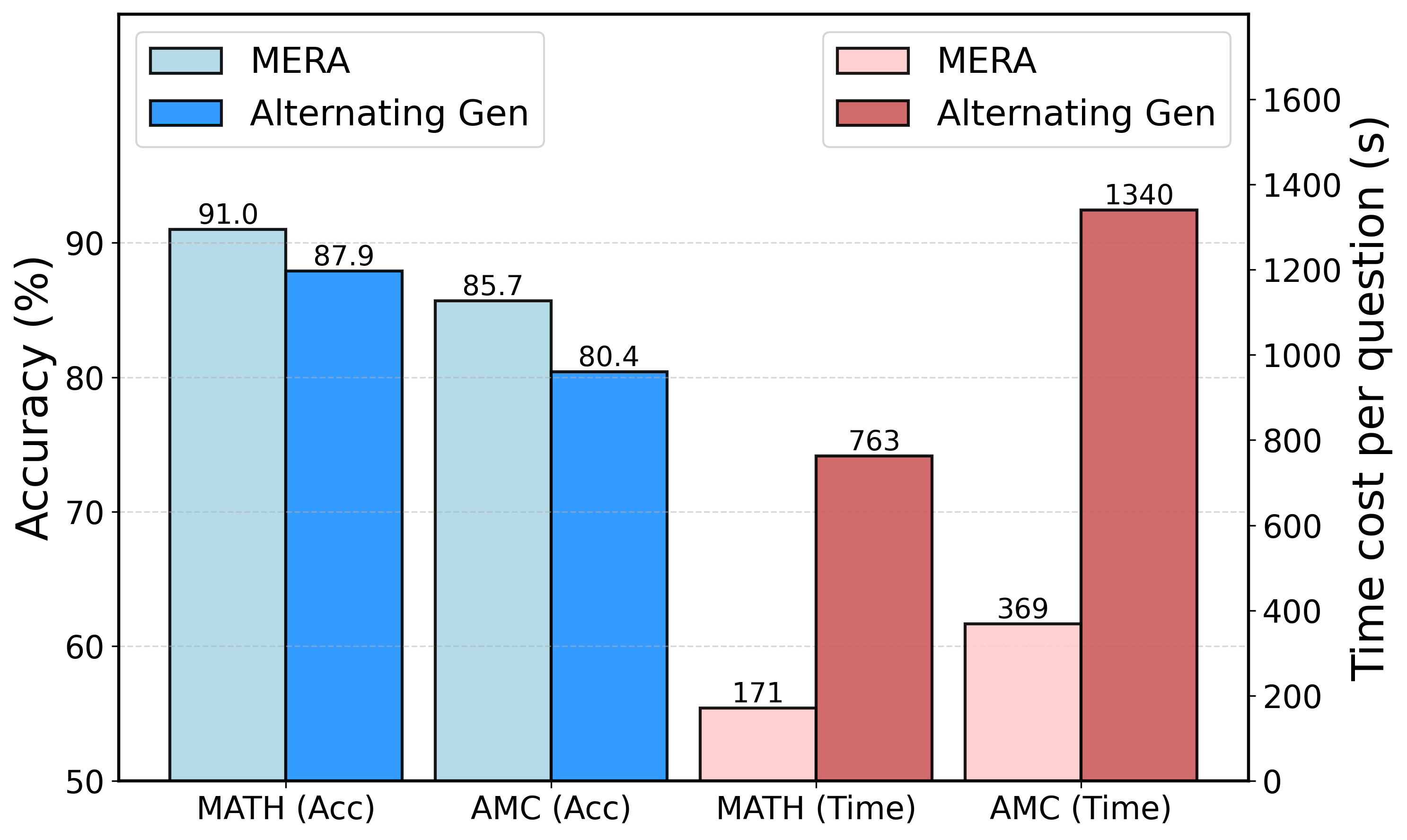}
    \caption{Comparison between MERA and the alternating generation setting in terms of accuracy and runtime using DeepSeek-R1-Distill-Qwen-7B.}

   \label{fig:altgen}
\end{figure}

\subsubsection{Analysis of Control Behavior.}
To further examine the behavioral changes induced by MERA’s structured meta-cognitive design, we analyze control statement usage during reasoning, focusing on two aspects: the total number of control statements and the average length of each control statement. As shown in Figure~\ref{fig:control}, MERA significantly reduces the frequency of control statements across all evaluated datasets. For instance, on AIME 2024, the original model produces an average of 44 control sentences, whereas MERA reduces this number to just 17. This substantial decline indicates that, after explicit decoupling and independent reinforcement, the model learns to suppress unnecessary or ineffective cognitive behaviors, thereby promoting more concise and coherent reasoning trajectories. In contrast, the average length of individual control statements increases notably under MERA, rising from 15 tokens to 37 tokens on AIME 2024. This trend becomes more pronounced as task difficulty increases. This suggests that the optimized control statements produced by the model tend to be more informative and carry stronger regulatory intent. Overall, these results demonstrate that MERA not only reduces redundant control interventions but also enhances the quality and functional value of retained control, enabling more deliberate and efficient self-regulation.

\begin{table*}[t]
\centering
\label{table2}
\setlength\tabcolsep{1pt} 
\begin{tabular}{@{}lccccccccccll@{}} 
\toprule
 \multirow{2}{*}{\textbf{Method}} 
 & \multicolumn{2}{c}{\textbf{GSM8K}} & \multicolumn{2}{c}{\textbf{MATH-500}} & \multicolumn{2}{c}{\textbf{AMC 2023}} & \multicolumn{2}{c}{\textbf{AIME 2024}} & \multicolumn{2}{c}{\textbf{AIME 2025}}    & \multicolumn{2}{|c}{\textbf{Overall}} \\
    & {Acc$\uparrow$} & {Tokens$\downarrow$} & {Acc$\uparrow$} & {Tokens$\downarrow$} & {Acc$\uparrow$}  & {Tokens$\downarrow$} & {Acc$\uparrow$} & {Tokens$\downarrow$} & {Acc$\uparrow$} & {Tokens$\downarrow$}  & {Acc$\uparrow$} & {Tokens$\downarrow$} \\ 
\hline
\multicolumn{13}{l}{{\cellcolor[cmyk]{0.017,0.017,0.017,0.017}}\textit{\textbf{DeepSeek-R1-Distill-Qwen-7B}}} \\
\textit{Original} & 90.2 & 1,819 & 86.9 & 3,422 & 77.4 & 6,738 & 53.1 & 12,185 & 48.2 & 13,276 &  \multicolumn{1}{|l}{71.2} & 7,488 \\ 

\textit{+SFT} & {91.2} & 1,575 & {88.1} & 2,340 & {80.5} & 4,881 & {53.6} & 9,171 & {52.7} & 9,934  & \multicolumn{1}{|l}{73.2} & 5,580  \\
\textit{+SFT+GRPO} & {91.9} & 1,785 & {89.5} & 2,614 & 83.7 & 5,193 & {55.4} & 9,740 & {52.9} & 11,059  & \multicolumn{1}{|l}{74.7} & 6,078  \\
\textit{+total MERA} & \textbf{93.7} & \textbf{822} & \textbf{91.0} & \textbf{1,739} & \textbf{85.7} & \textbf{3,711} & \textbf{56.1} & \textbf{8,398} & \textbf{53.6} & \textbf{8,732} & \multicolumn{1}{|l}{\textbf{76.0}} & \textbf{4,680}  \\
 \bottomrule
\end{tabular}
\caption{Ablation studies comparing different training strategies using DeepSeek-R1-Distill-Qwen-7B.}
\label{tab:ablation}
\end{table*}

\subsubsection{Generalization to different domains.} 
As shown in Table~\ref{tbl:diff_domains}, MERA achieves the highest reasoning efficiency across all three domains in the MMLU-Pro benchmark: Law, Engineering, and Physics, while maintaining accuracy on par with baseline methods. Specifically, MERA reduces the average token length by a large margin, such as from 5,412 to 4,092 in the Engineering, without sacrificing answer correctness. This indicates that the proposed meta-cognitive framework not only improves reasoning within mathematical tasks but also generalizes effectively to broader open-domain contexts. The consistent length reduction suggests the model can monitor its internal process and reduce redundancy, even on non-mathematical tasks.

\subsubsection{Comparison Between Internalized and Alternating Control Generation.} 
To evaluate the runtime efficiency and effectiveness of MERA’s internalized control mechanism, we compare it against a dual-model alternating generation setting, where a reasoning model and a separate auxiliary model jointly generate reasoning and control segments in turn. This setup mirrors the process used during data construction, in which control decisions are externally injected by a helper model. As shown in Figure~\ref{fig:altgen}, the alternating generation strategy results in significantly higher time cost per question, reaching 763 seconds on MATH and 1,340 seconds on AMC 2023, compared to MERA’s 171 and 369 seconds respectively. Moreover, MERA also achieves higher accuracy, outperforming the alternating method by 3.1\% on MATH and 5.3\% on AMC 2023. These results highlight the limitations of external control reliance at inference time. While such methods may provide high-quality control during data annotation, they introduce substantial latency and fail to match the coordination quality of a model with internalized control. In contrast, MERA achieves better efficiency and accuracy by integrating control and reasoning within a unified architecture, enabling coherent and cost-effective decisions.

\subsubsection{Ablation Study} 
Table~\ref{tab:ablation} presents the ablation study results. The results indicate that incorporating supervised SFT leads to a modest improvement in accuracy across all benchmark tasks, while also reducing token usage. When SFT is combined with standard GRPO, a slight increase in accuracy is observed, but at the cost of increased token generation. In contrast, the most substantial performance gains are achieved when the full MERA is applied. It not only yields higher accuracy but also enables more efficient reasoning. This shows that CSPO outperforms standard GRPO by optimizing control behavior.

\section{Related Work}
To mitigate overthinking in Large Reasoning Models (LRMs), prior work mainly follows three strategies. First, some methods directly compress reasoning trajectories: O1-Pruner~\cite{o1-pruner} trims LLM-generated traces for supervised distillation, and No Wait~\cite{nowait} suppresses hesitation tokens to shorten generation. While effective, such compression is often task-insensitive and can over-truncate. Second, budget-based controls constrain inference with predefined limits; for example, LCPO~\cite{aggarwal2025l1} sets a static token-length budget before decoding, but this coarse constraint cannot react to the model’s evolving reasoning state. Third, dynamic early-exit methods stop based on intermediate signals; DEER~\cite{yang2025dynamic} uses the confidence of partial answers to trigger termination, yet still relies on external evaluation rather than intrinsic self-monitoring. 

In contrast, our proposed MERA explicitly separates reasoning and control within a structured framework and optimizes them independently. MERA enables the model to autonomously decide whether to continue, revise, or terminate the reasoning process. This intrinsic regulatory capability allows for more adaptive and efficient reasoning by reducing unnecessary steps without relying on external heuristics or predefined constraints.
Additional related work is provided in the appendix~\ref{appendix:related}.

\section{Conclusion}
We propose MERA, a meta-cognitive framework that equips LRMs with structured control to mitigate overthinking. MERA separates reasoning from control, constructs high-quality control supervision via a takeover mechanism, and improves adaptive regulation with CSPO. Experiments show that MERA reduces redundant reasoning while improving accuracy across diverse benchmarks.

\section*{Limitations} 
MERA delivers consistent gains in both efficiency and accuracy, but two aspects warrant further improvement. First, support for user-side preference personalization remains limited, as different application contexts and users may favor different trade-offs in reasoning depth, verification intensity, and early-stopping aggressiveness; providing explicit preference controls would better tailor the control policy to these needs. Second, control-decision interpretability still has room for improvement: while the model can provide some rationale for actions such as stopping, continuing, or backtracking, scenarios that require higher transparency would benefit from making these rationales more fine-grained and auditable, and more explicitly tied to the key evidence and judgments in the reasoning process.

\section*{Acknowledgements}
This work is supported by the National Key Research and Development Program of China (2024YFF0907401), the National Natural ScienceFoundation of China (62072052) and Beijing Natural Science Foundation (L251037).

\normalem
\bibliography{custom}

\appendix

\section{Evaluation Dataset Descriptions}
\label{appendix:datasets}

We conduct experiments on a range of publicly available benchmark datasets. These datasets span various difficulty levels—from elementary arithmetic to Olympiad-style problems—and include cross-domain general knowledge evaluations. Detailed descriptions of each dataset are as follows:

\textbf{GSM8K}~\cite{data1} consists of approximately 8,500 high-quality elementary school math word problems designed to evaluate step-by-step numerical reasoning. Each problem typically involves multiple arithmetic operations and emphasizes accuracy, decomposition, and logical consistency.

\textbf{MATH-500}~\cite{data3} is a curated subset of 500 challenging problems sampled from the full MATH dataset. It covers a wide range of topics in high school and early college mathematics, including algebra, combinatorics, geometry, and number theory. MATH-500 is widely used to assess formal mathematical reasoning, requiring symbolic manipulation, inductive strategies, and structured derivations. It is particularly suited to evaluating the rigor and reliability of multi-hop mathematical.

\textbf{AMC2023} refers to the complete set of 40 problems from the 2023 American Mathematics Competition (AMC). The problems range from moderate to high difficulty and test creative decomposition, numerical estimation, and structural understanding. All problems are converted to a free-form generation format to align with our evaluation framework.

\textbf{AIME.}~\cite{data8} We collect problems from the 2024 and 2025 editions of the AIME (American Invitational Mathematics Examination), forming two small but highly challenging evaluation sets. Compared to AMC, AIME problems exhibit higher complexity and frequently involve algebraic constructions, advanced factorization techniques, and recursive formula analysis. 

\textbf{MMLU-Pro.}~\cite{data9} To evaluate generalization beyond mathematical domains, we use a subset of MMLU-Pro as our cross-domain benchmark. We select 300 problems in total—100 each from the domains of Physics, Law, and Engineering. This dataset evaluates the model’s ability to transfer structured reasoning and control skills to non-mathematical contexts, testing the robustness and broad applicability of our method.

\section{Algorithm}
\textbf{Algorithm \ref{algo:cspo}} outlines the Control Segment Policy Optimization (CSPO) procedure. The model first samples multiple reasoning–control trajectories and partitions them into segments. Each segment receives a control-specific reward based on semantic alignment and structural correctness. A masking mechanism ensures that only control-relevant tokens influence gradient updates. Finally, the policy is optimized using masked advantage-weighted updates, enabling more precise and stable learning of meta-cognitive control.

\begin{algorithm}[H]
\caption{Control Segment Policy Optimization}
\label{algo:cspo}
\begin{algorithmic}[1]
\Require Input-query $x$, Current policy $\pi_\theta$, Old policy $\pi_{\theta_{\text{old}}}$
\Ensure Updated policy with improved control precision

\State \textbf{Step 1: Segment Sampling}
\State Sample $G$ complete outputs $\{o_1, o_2, \dots, o_G\}$ from $\pi_{\theta_{\text{old}}}$
\State Partition each output into reasoning–control segments $\{\tau_1, \dots, \tau_K\}$

\State \textbf{Step 2: Segment-wise Reward Estimation}
\For{each segment $o_i$}
    \State Compute control reward $r(o_i)$ based on semantic and format signals
    \State Normalize across samples to obtain $\hat{A}_{i,t}$
\EndFor

\State \textbf{Step 3: Control Masking}
\State Identify control-relevant spans $\{c_1, \dots, c_K\}$ and construct mask $M_k$

\State \textbf{Step 4: Objective Update}
\For{each control token $t$ in $o_i$}
    \If{$M_k = 1$}
        \State Accumulate policy into $\mathcal{J}_{\text{CSPO}}(\theta)$
    \EndIf
\EndFor

\State \textbf{Step 5: Policy Optimization}
\State Update parameters via gradient descent on $\mathcal{J}_{\text{CSPO}}(\theta)$

\end{algorithmic}
\end{algorithm}

\section{Preliminary: Group Relative Policy Optimization (GRPO)}
\label{appendix:grpo}

Group Relative Policy Optimization (GRPO)~\cite{guo2025deepseek} is a reinforcement learning algorithm that estimates token-level advantages by comparing sampled sequences within the same input group. Unlike traditional methods such as PPO, GRPO avoids the use of a separate value function and instead derives advantage estimates based solely on normalized group-level rewards.

Given an input \(x\), the reference policy \(\pi_{\theta_{\text{old}}}\) generates a set of \(G\) candidate outputs \(\{y^{(i)}\}_{i=1}^{G}\). For each response \(y^{(i)}\), we compute a final reward \(\mathcal{R}(x, y^{(i)})\), and define the group-normalized advantage \(\hat{A}_i\) as:
\begin{equation}
\hat{A}_i = \frac{\mathcal{R}(x, y^{(i)}) - \text{mean}(\{\mathcal{R}(x, y^{(j)})\}_{j=1}^{G})}{\text{std}(\{\mathcal{R}(x, y^{(j)})\}_{j=1}^{G})}.
\end{equation}

This scalar advantage \(\hat{A}_i\) is uniformly assigned to each token in trajectory \(y^{(i)}\), such that \(\hat{A}_{i,t} = \hat{A}_i\) for all \(t \in \{1, \dots, |y^{(i)}|\}\).

The overall GRPO training objective incorporates a clipped policy loss and a KL regularization term with respect to a reference policy \(\pi_{\text{ref}}\), and is formulated as:

\begin{equation}\small
\begin{aligned}
\mathcal{J}_{\text{GRPO}}(\theta) =\ & \mathbb{E}_{x \sim \mathcal{D},\ \{y^{(i)}\}_{i=1}^{G} \sim \pi_\theta(\cdot|x)} \Bigg\{ 
\frac{1}{G} \sum_{i=1}^{G} \frac{1}{|y^{(i)}|} \sum_{t=1}^{|y^{(i)}|} \Big[ \\
& \min \big( r_{i,t}(\theta)\ \hat{A}_{i,t},\ \text{clip}(r_{i,t}(\theta), 1 - \epsilon,\ 1 + \epsilon)\  \\
&\hat{A}_{i,t} \big) - \beta\ D_{\text{KL}}(\pi_\theta \parallel \pi_{\text{ref}}) \Big] \Bigg\},
\end{aligned}
\end{equation}

where the policy ratio \(r_{i,t}(\theta)\) is given by:
\begin{equation}
r_{i,t}(\theta) = 
\frac{
\pi_\theta(a_{i,t} \mid s_{i,t})
}{
\pi_{\theta_{\text{old}}}(a_{i,t} \mid s_{i,t})
}.
\end{equation}

This token-level formulation supports trajectory-wise normalization while preserving fine-grained optimization, making GRPO well-suited for open-ended sequence generation scenarios such as mathematical reasoning and control-sensitive tasks.

\section{Extended Discussion on Related Work}
\label{appendix:related}

\textbf{Mitigating Overthinking in Large Reasoning Models (LRMs).}  
In complex multi-step tasks such as mathematical reasoning, Large Reasoning Models (LRMs) often exhibit excessively long, repetitive, or inefficient reasoning trajectories—a phenomenon referred to as "overthinking." Prior efforts to address this issue can be categorized into three major paradigms: (1) Direct trajectory shortening, (2) Budget-aware inference control, and (3) Output-based dynamic early termination.

\subsection*{Direct Trajectory Shortening}  
This class of methods seeks to compress existing reasoning trajectories by removing redundant steps while preserving correctness. For instance, O1-Pruner~\cite{o1-pruner} selects supervision signals from LLM-generated reasoning trajectories using a length-harmonized score that balances correctness and brevity. No Wait~\cite{nowait} reduces the probability of generating delay-inducing tokens such as “wait,” while No Think~\cite{nothink} directly instructs the model to skip intermediate reasoning and provide an answer immediately. Although effective at reducing output length, these approaches often lack adaptivity to problem complexity. Their uniform truncation strategies tend to harm performance on harder, more demanding problems.

\subsection*{Budget-Aware Inference Control}  
This line of work imposes length constraints or computational budgets before inference begins. For example, DAST~\cite{dast} applies fixed token-length budgets during preference learning, encouraging shorter outputs for correct responses and longer ones for incorrect cases. FCS+Ref~\cite{overthinking} constructs preference pairs based on identifying the first correct solution and reflection point within the reasoning trace. LCPO~\cite{aggarwal2025l1} adopts a straightforward reinforcement learning framework that jointly optimizes accuracy and user-specified length bounds. Moreover, Ada-Bok~\cite{adabok}, Thinkless~\cite{thinkless}, HGPO~\cite{hgpo} explore coarse-grained “fast–slow thinking” modes by dynamically deciding whether to engage in full CoT reasoning based on problem difficulty. A common limitation of these methods is that their control signals are predefined and fixed prior to generation, lacking responsiveness to the model’s evolving internal reasoning state.

\subsection*{Dynamic Early Exit}  
The third category monitors intermediate outputs during inference to dynamically determine when to terminate reasoning. DEER~\cite{yang2025dynamic} is a representative method in this direction, triggering early exits based on confidence scores of partial answers to improve efficiency while maintaining accuracy. Other methods leverage uncertainty signals such as entropy. For example, Entropy-Based Adaptive Think~\cite{entropythink} stops the generation process once marginal information gain drops below a threshold, avoiding inefficient stagnation. While these approaches introduce elements of self-regulation, they primarily rely on external scorers or auxiliary prediction heads and lack true meta-cognitive control within the model itself.

In contrast to the above paradigms, our proposed MERA framework explicitly separates reasoning and control within a structured modeling framework. MERA endows the model with an intrinsic self-regulatory mechanism that enables it to determine whether to continue, revise, or terminate its reasoning trajectory. Instead of relying on compression, budget, or external signals, MERA monitors internal reasoning states and inserts explicit \texttt{<control>} directives. These serve as fine-grained, adaptive interventions, enabling meta-cognitive regulation. Our training pipeline combines supervised fine-tuning with segment-level policy optimization (CSPO), allowing the model to develop both expressive reasoning capabilities and dynamic self-control.

\label{sec:appendix}

\end{document}